\documentclass[11pt,a4paper]{article}

\usepackage[hyperref]{emnlp-ijcnlp-2019}
\usepackage{times}
\usepackage{booktabs}
\usepackage{latexsym}

\usepackage{float}
\usepackage{hyperref}
\usepackage{graphicx}
\usepackage{makecell}
\usepackage{wrapfig}
\usepackage{url}

\aclfinalcopy 

\title{Neural Text Summarization: A Critical Evaluation}

\author{Wojciech Kry\'sci\'nski, Nitish Shirish Keskar, Bryan McCann, \\ 
        \textbf{Caiming Xiong, Richard Socher} \\
        Salesforce Research \\
        \texttt{\{kryscinski,nkeskar,bmccann,cxiong,rsocher\}@salesforce.com} \\}
  
\date{}
\begin{document}

\maketitle
\begin{abstract}
Text summarization aims at compressing long documents into a shorter form that conveys the most important parts of the original document.
Despite increased interest in the community and notable research effort, progress on benchmark datasets has stagnated.
We critically evaluate key ingredients of the current  research  setup:  datasets,  evaluation  metrics,  and  models, and   
highlight three primary shortcomings:
1)~automatically collected datasets leave the task underconstrained and may contain noise detrimental to training and evaluation, 
2)~current evaluation protocol is weakly correlated with human judgment and does not account for important characteristics such as factual correctness,
3)~models overfit to layout biases of current datasets and offer limited diversity in their outputs.
\end{abstract}
\section{Introduction} \label{sec:introduction}
Text summarization aims at compressing long textual documents into a short, human readable form that contains the most important information from the source.
Two strategies of generating summaries are \textit{extractive} \citep{Dorr:03, Nallapati:17}, where salient fragments of the source document are identified and directly copied into the summary, and \textit{abstractive} \citep{Rush:15, See:17}, where the salient parts are detected and paraphrased to form the final output.

The number of summarization models introduced every year has been increasing rapidly.
Advancements in neural network architectures~\cite{Sutskever:14, Bahdanau:14, Vinyals:15, Vaswani:17} and the availability of large scale data \cite{Sandhaus:08, Nallapati:16, Grusky:18} enabled the transition from systems based on expert knowledge and heuristics to data-driven approaches powered by end-to-end deep neural models.
Current approaches to text summarization utilize advanced attention and copying mechanisms \citep{See:17, Tan:17, Cohan:18}, multi-task and multi-reward training techniques~\cite{Guo:18, Pasunuru:18, Kryscinski:18}, reinforcement learning strategies~\citep{Paulus:17, Narayan:18b, Dong:18, Wu:18}, and hybrid extractive-abstractive models~\citep{PLiu:18, Hsu:18, Gehrmann:18, Chen:18}.
Many of the introduced models are trained on the CNN/DailyMail~\cite{Nallapati:16} news corpus, a popular benchmark for the field, and are evaluated based on $n$-gram overlap between the generated and target summaries with the ROUGE package~\citep{Lin:04}.

Despite substantial research effort, the progress on these benchmarks has stagnated.
State-of-the-art models only slightly outperform the Lead-3 baseline, which generates summaries by extracting the first three sentences of the source document. 
We argue that this stagnation can be partially attributed to the current research setup, which involves uncurated, automatically collected datasets and non-informative evaluations protocols.  We critically evaluate our hypothesis, and support our claims by analyzing three key components of the experimental setting: datasets, evaluation metrics, and model outputs.
Our motivation is to shift the focus of the research community into developing a more robust research setup for text summarization. 
\section{Related Work} \label{sec:related-work}
\subsection{Datasets}
To accommodate the requirements of modern data-driven approaches, several large-scale datasets have been proposed.
The majority of available corpora come from the news domain.
Gigaword~\citep{Graff:03} is a set of articles and corresponding titles that was originally used for headline generation~\citep{Takase:16}, but it has also been adapted to single-sentence summarization~\citep{Rush:15, Chopra:16}.
NYT~\citep{Sandhaus:08} is a collection of articles from the New York Times magazine with abstracts written by library scientists. It has been primarily used for extractive summarization \citep{Hong:14, Li:16} and phrase-importance prediction~\citep{Yang:14, Nye:15}.
The CNN/DailyMail \cite{Nallapati:16} dataset consists of articles with summaries composed of highlights from the article written by the authors themselves.
It is commonly used for both \textit{abstractive}~\citep{See:17, Paulus:17, Kryscinski:18} and \textit{extractive}~\citep{Dong:18, Wu:18, Zhao:18} neural summarization.
The collection was originally introduced as a Cloze-style QA dataset by~\citet{Hermann:15}.
XSum~\citep{Narayan:18} is a collection of articles associated with one, single-sentence summary targeted at abstractive models.
Newsroom~\citep{Grusky:18} is a diverse collection of articles sourced from 38 major online news outlets.
This dataset was released together with a leaderboard and held-out testing split.

Outside of the news domain, several datasets were collected from open discussion boards and other portals offering structure information.
Reddit TIFU~\citep{Kim:18} is a collection of posts scraped from Reddit where users post their daily stories and each post is required to contain a Too Long; Didn't Read (TL;DR) summary.
WikiHow~\citep{Koupaee:18} is a collection of articles from the WikiHow knowledge base, where each article contains instructions for performing procedural, multi-step tasks covering various areas, including: arts, finance, travel, and health. 

\subsection{Evaluation Metrics}
Manual and semi-automatic \citep{Nenkova:04, Passonneau:13} evaluation of large-scale summarization models is costly and cumbersome. 
Much effort has been made to develop automatic metrics that would allow for fast and cheap evaluation of models.

The ROUGE package~\citep{Lin:04} offers a set of automatic metrics based on the lexical overlap between candidate and reference summaries. 
Overlap can be computed between consecutive ($n$-grams) and non-consecutive (skip-grams) subsequences of tokens.
ROUGE scores are based on exact token matches, meaning that computing overlap between synonymous phrases is not supported.

Many approaches have extended ROUGE with support for synonyms and paraphrasing.
ParaEval~\citep{Zhou:06} uses a three-step comparison strategy, where the first two steps perform optimal and greedy paraphrase matching based on paraphrase tables before reverting to exact token overlap. 
ROUGE-WE~\citep{Ng:15} replaces exact lexical matches with a soft semantic similarity measure approximated with the cosine distances between distributed representations of tokens.
ROUGE 2.0~\citep{Ganesan:18} leverages synonym dictionaries, such as WordNet, and considers all synonyms of matched words when computing token overlap.
ROUGE-G~\citep{ShafieiBavani:18} combines lexical and semantic matching by applying graph analysis algorithms to the WordNet semantic network.
Despite being a step in the direction of a more comprehensive evaluation protocol, none of these metrics gained sufficient traction in the research community, leaving ROUGE as the default automatic evaluation toolkit for text summarization.

\subsection{Models}
Existing summarization models fall into three categories: \textit{abstractive}, \textit{extractive}, and \textit{hybrid}.

\textit{Extractive} models select spans of text from the input and copy them directly into the summary.
Non-neural approaches~\citep{Neto:02, Dorr:03, Filippova:13, Colmenares:15} utilized domain expertise to develop heuristics for summary content selection, whereas more recent, neural techniques allow for end-to-end training.
In the most common case, models are trained as word- or sentence-level classifiers that predict whether a fragment should be included in the summary~\citep{Nallapati:16b, Nallapati:17, Narayan:17, Liu:19, Xu:19}.
Other approaches apply reinforcement learning training strategies to directly optimize the model on task-specific, non-differentiable reward functions~\citep{Narayan:18b, Dong:18, Wu:18} .

\textit{Abstractive} models paraphrase the source documents and create summaries with novel phrases not present in the source document.
A common approach in \textit{abstractive} summarization is to use attention and copying mechanisms \citep{See:17, Tan:17, Cohan:18}.
Other approaches include using multi-task and multi-reward training~\cite{Paulus:17, Jiang:18, Guo:18, Pasunuru:18, Kryscinski:18}, and unsupervised training strategies \citep{Chu:18, Schumann:18}.

\textit{Hybrid} models~\citep{Hsu:18, PLiu:18, Gehrmann:18, Chen:18} include both \textit{extractive} and \textit{abstractive} modules and allow to separate the summarization process into two phases -- content selection and paraphrasing.

For the sake of brevity we do not describe details of different models, we refer interested readers to the original papers.

\subsection{Analysis and Critique}
Most summarization research revolves around new architectures and training strategies that improve the state of the art on benchmark problems.
However, it is also important to analyze and question the current methods and research settings.

\citet{Zhang:18} conducted a quantitative study of the level of abstraction in \textit{abstractive} summarization models and showed that word-level, copy-only \textit{extractive} models achieve comparable results to fully \textit{abstractive} models in the measured dimension. 
\citet{Kedzie:18}~offered a thorough analysis of how neural models perform content selection across different data domains, and exposed data biases that dominate the learning signal in the news domain and architectural limitations of current approaches in learning robust sentence-level representations.
\citet{Liu:10}~examine the correlation between ROUGE scores and human judgments when evaluating meeting summarization data and show that the correlation strength is low, but can be improved by leveraging unique meeting characteristics, such as available speaker information.
\citet{Owczarzak:12}~inspect how inconsistencies in human annotator judgments affect the ranking of summaries and correlations with automatic evaluation metrics. The results showed that system-level rankings, considering all summaries, were stable despite inconsistencies in judgments, however, summary-level rankings and automatic metric correlations benefit from improving annotator consistency.
\citet{Graham:15}~compare the fitness of the BLEU metric~\citep{Papineni:02} and a number of different ROUGE variants for evaluating summarization outputs. The study reveals superior variants of ROUGE that are different from the commonly used recommendations and shows that the BLEU metric achieves strong correlations with human assessments of generated summaries.
\citet{Schulman:15}~study the problems related to using ROUGE as an evaluation metric with respect to finding optimal solutions and provide proof of NP-hardness of global optimization with respect to ROUGE.

Similar lines of research, where the authors put under scrutiny existing methodologies, datasets, or models were conducted by \citet{CallisonBurch:06, CallisonBurch:07, Tan:15, Post:18} in machine translation, \cite{Gkatzia:15, Reiter:09, Reiter:18} in natural language generation, \citet{Lee:15, ChenB:16, Kaushik:18} in reading comprehension, \citet{Gururangan:18, Poliak:18, Glockner:18} in natural language inference, \citet{Goyal:17} in visual question answering, and \citet{Xian:17} in zero-shot image classification.
Comments on the general state of scholarship in the field of machine learning were presented by \citet{Sculley:18, Lipton:19} and references therein.
\section{Datasets} \label{sec:datasets}
\subsection{Underconstrained task}
\begin{table*}[th]
    \begin{center}
    \resizebox{\linewidth}{!}{%
    \small
    \begin{tabular}{p{0.48\linewidth}p{0.48\linewidth}} 
    \toprule
    \textbf{Article} \\ 
    \midrule
    \multicolumn{2}{p{\linewidth}}{The glowing blue letters that once lit the Bronx from above Yankee stadium failed to find a buyer at \textcolor{orange}{an auction at Sotheby's on Wednesday}. While the 13 letters were expected to bring in anywhere from \$300,000 to \$600,000, the only person who raised a paddle - for \$260,000 - was a Sotheby's employee trying to jump start the bidding. The \textcolor{magenta}{current owner of the signage is Yankee hall-of-famer Reggie Jackson}, \textcolor{blue}{who purchased the 10-feet-tall letters for an undisclosed amount after the stadium saw its final game in 2008}.  No love: 13 letters that hung over Yankee stadium were estimated to bring in anywhere from \$300,000 to \$600,000, but received no bids at a Sotheby's auction Wednesday. The 68-year-old Yankee said he wanted 'a new generation to own and enjoy this icon of the Yankees and of New York City.', The \textcolor{blue}{letters had beamed from atop Yankee stadium near grand concourse in the Bronx since 1976}, the year before Jackson joined the team. (...)} \\
    \midrule
    \textbf{}  \textbf{Summary Questions} & \\ 
    \midrule
    \makecell[l]{\textcolor{orange}{When was the auction at Sotheby's?} \\ \textcolor{magenta}{Who is the owner of the signage?} \\ \textcolor{blue}{When had the letters been installed on the stadium?}}  &\\
    
    \midrule
    \textbf{Constrained Summary A} & \textbf{Unconstrained Summary A}  \\ 
    \midrule
    Glowing letters that \textcolor{blue}{had been hanging above the Yankee stadium from 1976 to 2008} were placed for \textcolor{orange}{auction at Sotheby's on Wednesday}, but were not sold, The \textcolor{magenta}{current owner of the sign is Reggie Jackson}, a Yankee hall-of-famer. 
    & 
    There was not a single buyer at the \textcolor{orange}{auction at Sotheby's on Wednesday} for the glowing blue letters that once lit the Bronx's Yankee Stadium. Not a single non-employee raised their paddle to bid. \textcolor{magenta}{Jackson, the owner of the letters}, was surprised by the lack of results. The venue is also auctioning off other items like Mets memorabilia.\\
    \midrule
    \textbf{Constrained Summary B} & \textbf{Unconstrained Summary B} \\ 
    \midrule
    An auction for the lights from Yankee Stadium failed to \textcolor{orange}{produce any bids on Wednesday at Sotheby's}. The lights, \textcolor{magenta}{currently owned by former Yankees player Reggie Jackson}, \textcolor{blue}{lit the stadium from 1976 until 2008}. 
    &
    The once iconic and attractive pack of 13 letters that was \textcolor{orange}{placed at the Yankee stadium in 1976 and later removed in 2008} was unexpectedly not favorably considered at the Sotheby's auction when the 68 year old owner of the letters attempted to transfer its ownership to a member the younger populace. Thus, when the minimum estimate of \$300,000 was not met, a further attempt was made by a former player of the Yankees to personally visit the new owner as an
    \\
    \bottomrule
    \end{tabular}
    }%
    \caption{
    Example summaries collected from human annotators in the \textit{constrained} (left) and \textit{unconstrained} (right) task.
    In the \textit{unconstrained} setting, annotators were given a news article and asked to write a summary covering the parts they considered most important.
    In the \textit{constrained} setting, annotators were given  a news article with three associated questions and asked to write a summary that contained the answers to the given questions.
    }
    \label{tab:human-summaries}
    \end{center}
\end{table*}

The task of summarization is to compress long documents by identifying and extracting the most important information from the source documents.
However, assessing the importance of information is a difficult task in itself, that highly depends on the expectations and prior knowledge of the target reader.

We show that the current setting in which models are simply given a document with one associated reference summary and no additional information, leaves the task of summarization underconstrained and thus too ambiguous to be solved by end-to-end models.

To quantify this effect, we conducted a human study which measured the agreement between different annotators in selecting important sentences from a fragment of text.
We asked workers to write summaries of news articles and highlight sentences from the source documents that they based their summaries on.
The experiment was conducted in two settings:
\textit{unconstrained}, where the annotators were instructed to summarize the content that they considered most important, and
\textit{constrained}, where annotators were instructed to write summaries that would contain answers to three questions associated with each article. 
This is similar to the construction of the TAC 2008 Opinion Summarization Task \footnote{\url{https://tac.nist.gov/2008/summarization/op.summ.08.guidelines.html}}.
The questions associated with each article where collected from human workers through a separate assignment.
Experiments were conducted on 100 randomly sampled articles, further details of the human study can be found in Appendix~\ref{ssec:app-underconstrained-task}.

Table~\ref{tab:content-selection-agreement} shows the average number of sentences, per-article, that annotators agreed were important. 
The rows show how the average changes with the human vote threshold needed to reach consensus about the importance of any sentence.
For example, if we require that three or more human votes are necessary to consider a sentence important, annotators agreed on average on the importance of 0.627 and 1.392 sentences per article in the \textit{unconstrained} and \textit{constrained} settings respectively.
The average length (in sentences) of sampled articles was 16.59, with a standard deviation of 5.39.
The study demonstrates the difficulty and ambiguity of content selection in text summarization.

We also conducted a qualitative study of summaries written by annotators. 
Examples comparing summaries written in the \textit{constrained} and \textit{unconstrained} setting are shown in Table~\ref{tab:human-summaries}.
We noticed that in both cases the annotators correctly identified the main topic and important fragments of the source article.
However, \textit{constrained} summaries were more succinct and targeted, without sacrificing the natural flow of sentences.
\textit{Unconstrained} writers tended to write more verbose summaries that did not add information.
The study also highlights the \textit{abstractive} nature of human written summaries in that similar content can be described in unique ways.

\begin{table}[t!]
    \begin{center}
    \resizebox{\linewidth}{!}{%
        \small
        \begin{tabular}{c|cc} 
        \toprule
        & \multicolumn{2}{c}{Sent. per article considered important} \\
        \makecell{Human vote \\ threshold} & \textit{Unconstrained} & \textit{Constrained} \\ 
        \midrule
        $= 5$   & 0.028 & 0.251 \\ 
        $\ge 4$ & 0.213 & 0.712 \\ 
        $\ge 3$ & 0.627 & 1.392 \\ 
        $\ge 2$ & 1.695 & 2.404 \\ 
        $\ge 1$ & 5.413 & 4.524 \\ 
        \midrule
        \bottomrule
        \end{tabular}
    }%
    \caption{
    Average number of sentences, per-article, which annotators agreed were important.
    The human vote threshold investigates how the average agreement changes with the threshold of human votes required to consider any sentence important.
    Rows $= 5$ and $\ge 1$ correspond to the set intersection and union of selected sentences accordingly.
    }
    \label{tab:content-selection-agreement}
    \end{center}
\end{table}

\subsection{Layout bias in news data}
\begin{figure}[ht]
\centering
\includegraphics[width=\linewidth]{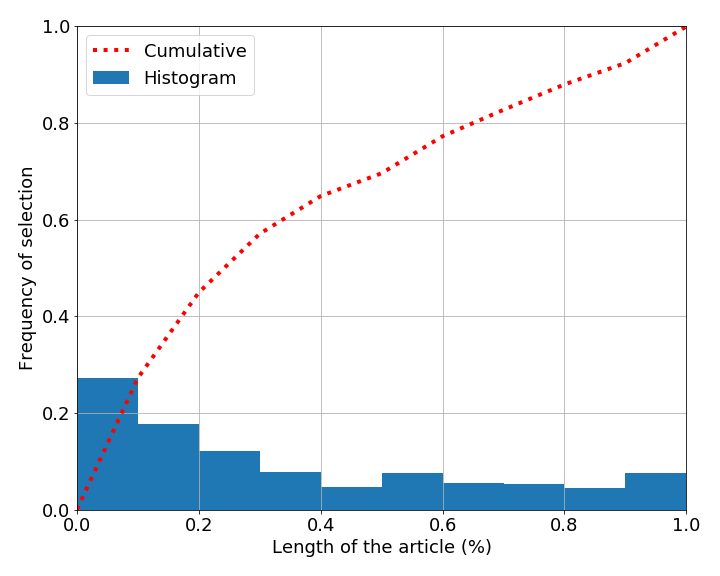}
\caption{
        The distribution of important sentences over the length of the article according to human annotators (blue) and its cumulative distribution (red).
        }
\label{fig:pyramid-bias}
\end{figure}

News articles adhere to a writing structure known in journalism as the "Inverted Pyramid"~\cite{Purdue:19}. 
In this form, initial paragraphs contain the most newsworthy information, which is followed by details and background information.

To quantify how strongly articles in the CNN/DM corpus follow this pattern we conducted a human study that measured the importance of different sections of the article.
Annotators read news articles and selected sentences they found most important. 
Experiments were conducted on 100 randomly sampled articles, further details of the human study are described in Appendix~\ref{ssec:app-layout-bias}.
Figure~\ref{fig:pyramid-bias} presents how annotator selections were distributed over the length of the article. 
The distribution is skewed towards the first quarter of the length of articles. 
The cumulative plot shows that nearly 60\% of the important information was present in the first third of the article, and approximately 25\% and 15\% of selections pointing to the second and last third, respectively.

It has become standard practice to exploit such biases during training to increase performance of models ~\cite{See:17, Paulus:17, Kryscinski:18, Gehrmann:18, Jiang:18, Pasunuru:18}, but the importance of these heuristics has been accepted without being quantified.
These same heuristics would not apply to books or legal documents, which lack the Inverted Pyramid layout so common in the news domain,
so it is important that these heuristics be part of ablation studies rather than accepted as default pre-processing step.

\subsection{Noise in scraped datasets}
Given the data requirements of deep neural networks and the vast amounts of diverse resources available online, automatically scraping web content is a convenient way of collecting data for new corpora.
However, adapting scraped content to the needs of end-to-end models is problematic.
Given that manual inspection of data is infeasible and human annotators are expensive, data curation is usually limited to removing any markup structure and applying simple heuristics to discard obviously flawed examples.
This, in turn, makes the quality of the datasets heavily dependent on how well the scraped content adheres to the assumptions made by the authors about its underlying structure.

This issue suggests that available summarization datasets would be filled with noisy examples.
Manual inspection of the data, particularly the reference summaries, revealed easily detectable, consistent patterns of flawed examples
Many such examples can be isolated using simple regular expressions and heuristics,
which allows approximation of how widespread these flaws are in the dataset.

We investigated this issue in two large summarization corpora scraped from the internet: CNN/DM \citep{Nallapati:16} and the Newsroom \citep{Grusky:18}.
The problem of noisy data affects 0.47\%, 5.92\%, and 4.19\% of the training, validation, and test split of the CNN/DM dataset, and 3.21\%, 3.22\%, and 3.17\% of the respective splits of the Newsroom dataset.
Examples of noisy summaries are shown in Table~\ref{tab:noisy-labels}.
Flawed examples contained links to other articles and news sources, placeholder texts, unparsed HTML code, and non-informative passages in the reference summaries.

\begin{table}[t]
    \begin{center}
    \resizebox{\linewidth}{!}{%
    \small
        \begin{tabular}{p{\linewidth}} 
        \toprule
        \textit{CNN/DM - Links to other articles} \\ 
        \midrule
        Michael Carrick has helped Manchester United win their last six games. 
        Carrick should be selected alongside Gary Cahill for England. 
        Carrick has been overlooked too many times by his country. 
        \textcolor{red}{READ : Carrick and Man United team-mates enjoy second Christmas party.} \\
        \midrule
        \textit{Newsroom - Links to news sources} \\ 
        \midrule
        \textcolor{red}{Get Washington DC, Virginia, Maryland and national news. 
        Get the latest/breaking news, featuring national security, science and courts. 
        Read news headlines from the nation and from The Washington Post. 
        Visit www.washingtonpost.com/nation today.} \\
        \bottomrule
        \end{tabular}
    }%
    \caption{
    Examples of noisy reference summaries found in the CNN/DM and Newsroom datasets.
    }
    \label{tab:noisy-labels}
    \end{center}
\end{table}

\begin{table}[t]
    \begin{center}
    \resizebox{\linewidth}{!}{%
    \small
        \begin{tabular}{p{\linewidth}} 
        \toprule
        \textit{Article} \\ 
        \midrule
        Quick-thinking: \textcolor{blue}{Brady Olson, a teacher at North Thurston High, took down a gunman on Monday}. A Washington High School teacher is being hailed a hero for tackling a 16-year-old student to the ground after he opened fire on Monday morning (...) \\
        \midrule
        \textit{Summary - Factually incorrect} \\
        \midrule
        \textcolor{red}{Brady Olson, a Washington High School teacher at North Thurston High, opened fire on Monday morning.} No one
        was injured after the boy shot twice toward the ceiling in the school commons before classes began at North Thurston High School in Lacey (...) \\
        \bottomrule
        \end{tabular}
        }%
    \caption{
    Example of a factually incorrect summary generated by an abstractive model. Top: ground-truth article. Bottom: summary generated by model.
    }
    \label{tab:factual-errors}
    \end{center}
\end{table}
\section{Evaluation Metrics} \label{sec:evaluation-metrics}
\subsection{Weak correlation with human judgment}
\begin{table*}[th!]
    \begin{center}
    \resizebox{\linewidth}{!}{%
    \small
        \begin{tabular}{r|ccc|ccc|ccc|ccc|ccc|ccc|} 
        & \multicolumn{9}{c|}{Pearson correlation} & \multicolumn{9}{c|}{Kendall rank correlation} \\
        & \multicolumn{3}{c|}{1 Reference} & \multicolumn{3}{c|}{5 References} & \multicolumn{3}{c|}{10 References} 
        & \multicolumn{3}{c|}{1 Reference} & \multicolumn{3}{c|}{5 References} & \multicolumn{3}{c|}{10 References} \\
        & \textbf{R-1} & \textbf{R-2} & \textbf{R-L} & \textbf{R-1} & \textbf{R-2} & \textbf{R-L} 
        & \textbf{R-1} & \textbf{R-2} & \textbf{R-L} & \textbf{R-1} & \textbf{R-2} & \textbf{R-L}
        & \textbf{R-1} & \textbf{R-2} & \textbf{R-L} & \textbf{R-1} & \textbf{R-2} & \textbf{R-L} \\
        \midrule
        \multicolumn{19}{c}{\textit{All Models}} \\
        \midrule
        Relevance   & $0.07$ & $0.03$ & $0.06$ & $0.03$ & $0.02$ & $0.02$ & $0.05$ & $0.03$ & $0.04$ 
                    & $0.29$ & $0.30$ & $0.29$ & $0.28$ & $0.29$ & $0.27$ & $0.28$ & $0.28$ & $0.27$ \\
        Consistency & $0.08$ & $0.03$ & $0.07$ & $0.02$ & $0.01$ & $0.01$ & $0.03$ & $0.01$ & $0.02$ 
                    & $0.27$ & $0.27$ & $0.28$ & $0.28$ & $0.28$ & $0.27$ & $0.28$ & $0.30$ & $0.28$ \\
        Fluency     & $0.08$ & $0.06$ & $0.08$ & $0.05$ & $0.03$ & $0.04$ & $0.05$ & $0.04$ & $0.05$
                    & $0.28$ & $0.27$ & $0.29$ & $0.26$ & $0.28$ & $0.27$ & $0.26$ & $0.26$ & $0.25$ \\ 
        Coherence   & $0.06$ & $0.05$ & $0.07$ & $0.05$ & $0.04$ & $0.05$ & $0.04$ & $0.03$ & $0.04$
                    & $0.29$ & $0.31$ & $0.29$ & $0.27$ & $0.27$ & $0.27$ & $0.28$ & $0.28$ & $0.27$ \\
        \midrule
        \multicolumn{19}{c}{\textit{Abstractive Models}} \\
        \midrule
        Relevance   & $0.04$ & $0.01$ & $0.05$ & $0.01$ & $0.00$  & $0.00$  & $0.04$ & $0.02$ & $0.03$ 
                    & $0.31$ & $0.28$ & $0.29$ & $0.25$ & $0.28$  & $0.27$  & $0.28$ & $0.26$ & $0.24$ \\
        Consistency & $0.07$ & $0.01$ & $0.06$ & $0.00$ & $-0.02$ & $-0.01$ & $0.03$ & $0.01$ & $0.03$ 
                    & $0.30$ & $0.30$ & $0.29$ & $0.26$ & $0.25$  & $0.25$  & $0.24$ & $0.27$ & $0.24$ \\
        Fluency     & $0.06$ & $0.04$ & $0.07$ & $0.03$ & $0.01$  & $0.02$  & $0.05$ & $0.04$ & $0.04$
                    & $0.28$ & $0.28$ & $0.28$ & $0.24$ & $0.25$  & $0.24$  & $0.27$ & $0.26$ & $0.24$ \\
        Coherence   & $0.04$ & $0.02$ & $0.04$ & $0.02$ & $0.01$  & $0.02$  & $0.03$ & $0.02$ & $0.03$
                    & $0.30$ & $0.32$ & $0.31$ & $0.25$ & $0.27$  & $0.26$  & $0.28$ & $0.29$ & $0.26$ \\
        \midrule
        \multicolumn{19}{c}{\textit{Extractive Models}} \\
        \midrule
        Relevance   & $0.14$ & $0.09$ & $0.13$ & $0.09$ & $0.05$ & $0.07$ & $0.06$ & $0.03$  & $0.04$ 
                    & $0.45$ & $0.44$ & $0.49$ & $0.38$ & $0.38$ & $0.38$ & $0.44$ & $0.43$  & $0.42$  \\
        Consistency & $0.10$ & $0.09$ & $0.11$ & $0.07$ & $0.07$ & $0.07$ & $0.00$ & $-0.03$ & $-0.02$ 
                    & $0.48$ & $0.49$ & $0.49$ & $0.38$ & $0.39$ & $0.38$ & $0.42$ & $0.41$  & $0.42$  \\
        Fluency     & $0.13$ & $0.14$ & $0.13$ & $0.10$ & $0.10$ & $0.08$ & $0.06$ & $0.03$  & $0.04$
                    & $0.47$ & $0.47$ & $0.48$ & $0.35$ & $0.37$ & $0.35$ & $0.39$ & $0.34$  & $0.37$  \\
        Coherence   & $0.15$ & $0.17$ & $0.15$ & $0.13$ & $0.13$ & $0.13$ & $0.08$ & $0.05$  & $0.06$
                    & $0.42$ & $0.43$ & $0.46$ & $0.40$ & $0.39$ & $0.39$ & $0.38$ & $0.34$  & $0.36$  \\
        \bottomrule
        \end{tabular}
        }%
    \caption{
    Correlations between human annotators and ROUGE scores along different dimensions and multiple reference set sizes.
    Left:~Pearson's correlation coefficients. 
    Right:~Kendall's rank correlation coefficients.
    }
    \label{tab:rouge-correlations}
    \end{center}
\end{table*}

The effectiveness of ROUGE was previously evaluated~\cite{Lin:04, Graham:15} through statistical correlations with human judgment on the DUC datasets~\citep{DUC:01, DUC:02, DUC:03}.
However, their setting was substantially different from the current environment in which summarization models are developed and evaluated.

To investigate the robustness of ROUGE in the setting in which it is currently used, we evaluate how its scores correlate with the judgment of an average English-speaker using examples from the CNN/DM dataset.
Following the human evaluation protocol from \citet{Gehrmann:18}, we asked annotators to rate summaries across four dimensions:
\textit{relevance} (selection of important content from the source),
\textit{consistency} (factual alignment between the summary and the source),
\textit{fluency} (quality of individual sentences), 
and \textit{coherence} (collective quality of all sentences).
Each summary was rated by 5 distinct judges with the final score obtained by averaging the individual scores.
Experiments were conducted on 100 randomly sampled articles with the outputs of 13 summarization systems provided by the original authors. 
Correlations were computed between all pairs of Human-, ROUGE-scores, for all systems.
Additional summaries were collected from annotators to inspect the effect of using multiple ground-truth labels on the correlation with automatic metrics.
Further details of the human study can be found in Appendix~\ref{ssec:app-rouge-correlation}.

Results are shown in Table~\ref{tab:rouge-correlations}. 
The left section of the table presents Pearson's correlation coefficients and the right section presents Kendall rank correlation coefficients.
In terms of Pearsons's coefficients, the study showed minimal correlation with any of the annotated dimensions for both \textit{abstractive} and \textit{extractive} models together and for \textit{abstractive} models individually.
Weak correlation was discovered for \textit{extractive} models primarily with the \textit{fluency} and \textit{coherence} dimensions.

We hypothesized that the noise contained in the fine-grained scores generated by both human annotators and ROUGE might have affected the correlation scores.
We evaluated the relation on a higher level of granularity by means of correlation between rankings of models that were obtained from the fine-grained scores.
The study showed weak correlation with all measured dimensions, when evaluated for both \textit{abstractive} and \textit{extractive} models together and for \textit{abstractive} models individually.
Moderate correlation was found for \textit{extractive} models across all dimensions.
A surprising result was that correlations grew weaker with the increase of ground truth references.

Our results align with the observations from \citet{Liu:10} who also evaluated ROUGE outside of its original setting.
The study highlights the limited utility in measuring progress of the field solely by means of ROUGE scores.

\subsection{Insufficient evaluation protocol}
The goal of text summarization is to automatically generate succinct, fluent, relevant, and factually consistent summaries.
The current evaluation protocol depends primarily on the exact lexical overlap between reference and candidate summaries measured by ROUGE.
In certain cases, ROUGE scores are complemented with human studies where annotators rate the \textit{relevance} and \textit{fluency} of generated summaries.
Neither of the methods explicitly examines the factual consistency of summaries, leaving this important dimension unchecked.

To evaluate the factual consistency of existing models, we manually inspected randomly sampled articles with summaries coming from randomly chosen, \textit{abstractive} models.
We focused exclusively on factual incorrectness and ignored any other issues, such as low fluency.
Out of 200 article-summary pairs that were reviewed manually, we found that 60 (30\%) contained consistency issues.
Table~\ref{tab:factual-errors} shows examples of discovered inconsistencies.
Some of the discovered inconsistencies, despite being factually incorrect, could be rationalized by humans.
However, in many cases, the errors were substantial and could have severe repercussions if presented as-is to target readers. 
\section{Models} \label{sec:models}

\begin{table*}[t]
    \begin{center}
    \resizebox{\linewidth}{!}{%
    \small
        \begin{tabular}{l|ccccc|ccccc} 
        & \multicolumn{5}{c|}{Target Reference} 
        & \multicolumn{5}{c}{Lead-3 Reference} \\
        & \textbf{R-1} & \textbf{R-2} & \textbf{R-3} & \textbf{R-4} &\textbf{R-L} 
        & \textbf{R-1} & \textbf{R-2} & \textbf{R-3} & \textbf{R-4} & \textbf{R-L} \\
        \midrule
        Extractive Oracle \cite{Grusky:18} & 93.36 & 83.19 & -    & -    & 93.36 
                                           & -     & -     & -    & -    & - \\ 
        Lead-3 Baseline                    & 40.24 & 17.53 & 9.94 & 6.50 & 36.49 
                                           & -     & -     & -    & -    & - \\ 
        \midrule
        \multicolumn{9}{c}{\textit{Abstractive Models}} \\
        \midrule
        Model \citet{Hsu:18}        & 40.68 & 17.97 & 10.43 & 6.97 & 37.13 
                                    & 69.66 & 62.60 & 60.33 & 58.72 & 68.42 \\ 
        Model \citet{Gehrmann:18}   & 41.53 & 18.77 & 10.68 & 6.98  & 38.39 
                                    & 52.25 & 39.03 & 33.40 & 29.61 & 50.21 \\
        Model \citet{Jiang:18}      & 40.05 & 17.66 & 10.34 & 6.99  & 36.73 
                                    & 62.32 & 52.93 & 49.95 & 47.98 & 60.72 \\
        Model \citet{Chen:18}       & 40.88 & 17.81 & 9.79  & 6.19  & 38.54 
                                    & 55.87 & 41.30 & 34.69 & 29.88 & 53.83 \\
        Model \citet{See:17}        & 39.53 & 17.29 & 10.05 & 6.75  & 36.39 
                                    & 58.15 & 47.60 & 44.11 & 41.82 & 56.34 \\ 
        Model \citet{Kryscinski:18} & 40.23 & 17.30 & 9.33  & 5.70  & 37.76 
                                    & 57.22 & 42.30 & 35.26 & 29.95 & 55.13 \\
        Model \citet{WLi:18}        & 40.78 & 17.70 & 9.76  & 6.19  & 38.34 
                                    & 56.45 & 42.36 & 35.97 & 31.39 & 54.51 \\ 
        Model \citet{Pasunuru:18}   & 40.44 & 18.03 & 10.56 & 7.12  & 37.02 
                                    & 62.81 & 53.57 & 50.25 & 47.99 & 61.27 \\ 
        Model \citet{Zhang:18}      & 39.75 & 17.32 & 10.11 & 6.83  & 36.54 
                                    & 58.82 & 47.55 & 44.07 & 41.84 & 56.83 \\ 
        Model \citet{Guo:18}        & 39.81 & 17.64 & 10.40 & 7.08  & 36.49 
                                    & 56.42 & 45.88 & 42.39 & 40.11 & 54.59 \\ 
        \midrule
        \multicolumn{9}{c}{\textit{Extractive Models}} \\
        \midrule
        Model \citet{Dong:18}      & 41.41 & 18.69 & 10.87 & 7.22  & 37.61 
                                   & 73.10 & 66.98 & 65.49 & 64.66 & 72.05 \\ 
        Model \citet{Wu:18}        & 41.25 & 18.87 & 11.05 & 7.38  & 37.75 
                                   & 78.68 & 74.74 & 73.74 & 73.12 & 78.08 \\
        Model \citet{Zhao:18}      & 41.59 & 19.00 & 11.13 & 7.45  & 38.08 
                                   & 69.32 & 61.00 & 58.51 & 56.98 & 67.85 \\ 
        \end{tabular}
        }%
    \caption{
    ROUGE (R-) scores computed for different models on the test set of the CNN/DM dataset.
    Left:~Scores computed with the original reference summaries.
    Right:~Scores computed with Lead-3 used as the reference.
    }
    \label{tab:lead-rouge}
    \end{center}
\end{table*}

\subsection{Layout bias in news data}
We revisit the problem of layout bias in news data from the perspective of models.
\citet{Kedzie:18}~showed that in the case of news articles, the layout bias dominates the learning signal for neural models.
In this section, we approximate the degree with which generated summaries rely on the leading sentences of news articles.

We computed ROUGE scores for collected models in two settings: first using the CNN/DM reference summaries as the ground-truth, and second where the leading three sentences of the source article were used as the ground-truth, i.e. the Lead-3 baseline. 
We present the results in Table~\ref{tab:lead-rouge}.

For all examined models we noticed a substantial increase of overlap across all ROUGE variants.
Results suggest that performance of current models is strongly affected by the layout bias of news corpora.
Lead-3 is a strong baseline that exploits the described layout bias.
However, there is still a large gap between its performance and an upper bound for extractive models (extractive oracle).

\subsection{Diversity of model outputs}
Models analyzed in this paper are considerably different from each other in terms of architectures, training strategies, and underlying approaches.
We inspected how the diversity in approaches translates into the diversity of model outputs.

\begin{figure}[ht!]
\centering
\includegraphics[width=\linewidth]{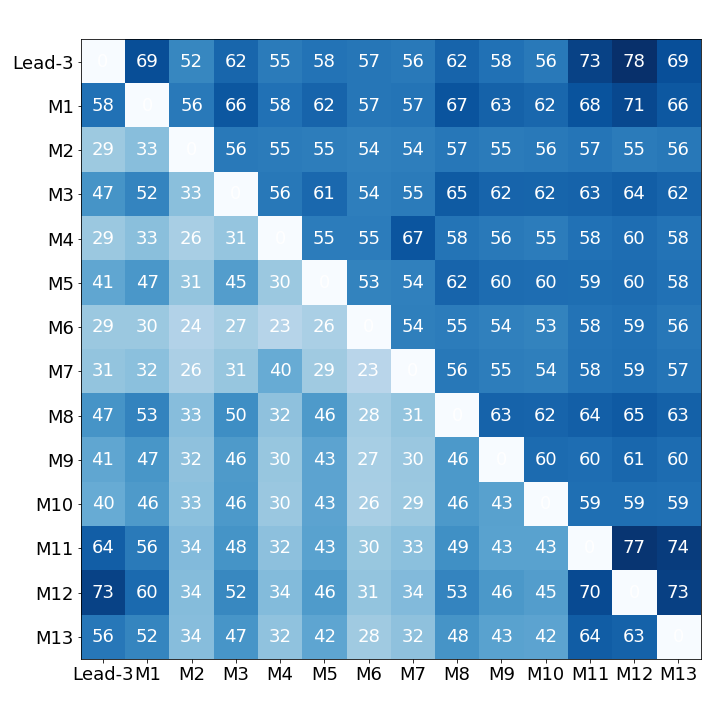}
\caption{
    Pairwise similarities between model outputs computed using ROUGE.
    Above diagonal: Unigram overlap (ROUGE-1).
    Below diagonal: 4-gram overlap (ROUGE-4).
    Model order (M-) follows Table~\ref{tab:lead-rouge}.
}
\label{fig:output-diversity}
\end{figure}

We computed ROUGE-1 and ROUGE-4 scores between pairs of model outputs to compare them by means of token and phrase overlap.
Results are visualized in Figure~\ref{fig:output-diversity}, where the values above and below the diagonal are ROUGE-1 and -4 scores accordingly, and model names (M-) follow the order from Table~\ref{tab:lead-rouge}.
\newpage
We notice that the ROUGE-1 scores vary considerably less than ROUGE-4 scores.
This suggests that the models share a large part of the vocabulary on the token level, but differ on how they organize the tokens into longer phrases.

Comparing results with the $n$-gram overlap between models and reference summaries (Table~\ref{tab:lead-rouge}) shows a substantially higher overlap between any model pair than between the models and reference summaries. 
This might imply that the training data contains easy to pick up patterns that all models overfit to, or that the information in the training signal is too weak to connect the content of the source articles with the reference summaries.

\section{Conclusions} \label{sec:conclusions}
This critique has highlighted the weak points of the current research setup in text summarization.
We showed that text summarization datasets require additional constraints to have well-formed summaries, current state-of-the-art methods learn to rely too heavily on layout bias associated with the particular domain of the text being summarized, and the current evaluation protocol reflects human judgments only weakly while also failing to evaluate critical features (e.g. factual correctness) of text summarization.

We hope that this critique provides the summarization community with practical insights for future research directions that include the construction of datasets, models less fit to a particular domain bias, and evaluation that goes beyond current metrics to capture the most important features of summarization.
\section{Acknowledgements} \label{sec:acknowledgements}
We thank all the authors listed in Table~\ref{tab:lead-rouge} for sharing their model outputs and thus contributing to this work. We also thank Shafiq Rayhan Joty for reviewing this manuscript and providing valuable feedback.

\clearpage
\bibliography{emnlp-ijcnlp-2019}
\bibliographystyle{acl_natbib}
\clearpage

\appendix
\section{Human study details}
Human studies were conducted through the Amazon Mechanical Turk platform.
Prices of tasks were carefully calculated to ensure that workers would have an average compensation of 12USD per hour.
In all studies, examples were sampled from the test split of the CNN/DM dataset that contains a total of 11,700 examples.

As with any human study, there is a trade-off between the number of examples annotated, the breadth of the experiments, and the quality of annotations.
Studies conducted for this paper were calibrated to primarily assure high quality of results and the breadth of experiments.

\subsection{Underconstrained task} \label{ssec:app-underconstrained-task}
Human annotators were asked to write summaries of news articles and highlight fragments of the source documents that they found useful for writing their summary.
The study was conducted on 100 randomly sampled articles, with each article annotated by 5 unique annotators.
The same configuration and articles were used in both the \textit{constrained} and \textit{unconstrained} setting.

Questions for the \textit{constrained} setting were written by human annotators in a separate assignment and curated before being used for to collect summaries.


\subsection{ROUGE - Weak correlation with human judgment} \label{ssec:app-rouge-correlation}
This study evaluated the quality of summaries generated by 13 different neural models, 10 \textit{abstractive} and 3 \textit{extractive}.
A list of evaluated models is available in Table~\ref{tab:lead-rouge}.

The study was conducted on 100 randomly sampled articles, with each article annotated by 5 unique annotators.
Given the large number of evaluated models, the experiment was split into 3 groups.
Two groups contained 4 models, one group contained 5 models.
To prevent from collecting biased data, models were assigned to experiment groups on a per-example basis, thus randomizing the context in which each model was evaluated.
To establish a common reference point between groups, the reference summaries from the dataset were added to the pool of annotated models, however, annotators were not informed which of this fact.
The order in which summaries were displayed in the annotation interface was randomized with the first position always reserved for the reference summary.

\subsection{Layout bias in news data} \label{ssec:app-layout-bias}
Human annotators were asked to read news articles and highlight the sentences that contained the most important information.
The study was conducted on 100 randomly sampled articles, with each article annotated by 5 unique annotators.

\clearpage
\end{document}